%% file: main.tex
\pdfoutput=1
\documentclass[11pt]{article}
\usepackage[final]{acl}
\usepackage{times}
\usepackage{latexsym}
\usepackage[T1]{fontenc}
\usepackage[utf8]{inputenc}
\usepackage{microtype}
\usepackage{inconsolata}

\usepackage{booktabs}
\usepackage{siunitx}
\usepackage{graphicx}
\usepackage{multirow}
\usepackage{xcolor}

\title{What the Window Does Not Contain: Auditing Provenance in a\\Document-Grounded Instability Benchmark}

\author{Seyed Mosayeb Alam \\
  KTH Royal Institute of Technology \\
  \texttt{smalam@kth.se}}

\makeatletter
\newcommand{\artifacturl}{https://github.com/eikiyo/FoFinNLP}
\newcommand{\artifactpointer}{%
  \ifx\artifacturl\@empty The artifact is at the anonymised URL in the submission form.%
  \else The artifact is at \url{\artifacturl}.\fi}
\makeatother

\begin{document}
\maketitle

\begin{abstract}
Ask a language model the same question about the same document twenty times, and it sometimes
returns two different answers. We built \textbf{Probity}, a benchmark of 60 tasks and 470 items from
real venture-financing filings, to measure how often this happens. Then we audited our own corpus
and found a defect any excerpt-built benchmark can carry: items whose evidence is missing from the
window of text the model is shown. The audit flags 36 items and separates two failures a single
flag would conflate: evidence genuinely absent from the window, and answers that must be computed
from numbers the window does supply. Flagged items change their answers far more often, wobbling at
0.255 against 0.087 on the 427 clean items, and excluding them cuts apparent cross-model agreement
by about a fifth. Before testing whether the missing evidence explains the instability, we
registered a prediction: re-cut each window to hold its evidence, and instability should fall below
a set threshold. It failed: the repair moved wobble by 0.058, with an interval containing zero. We
report the association as correlational. Almost all measurements sit where instability cannot show, which
bounds what a corpus built for accuracy can say about stability. We release the corpus, all
112{,}800 raw responses, and the audit as a runnable check for any document benchmark.
\end{abstract}

\section{Introduction}

Language models now read financing documents and answer questions people act on: what share of a
company an investor owns after a round, or who gets paid first, and how much, when the company is
sold. Ask twice and the number can differ. When it does, at least one answer is wrong, and nothing
marks which. An answer that changes between two identical calls reports a different
allocation of the same company.

We built Probity to measure that instability: 60 tasks over 470 items, each item one real filing (a
document a company files with the securities regulator) paired with one question about one clause.
Twelve model configurations each answer every item 20 times. We say an item wobbles when a
configuration does not return the same answer on all 20 runs.

Then we audited our own benchmark, and what the audit found is most of this paper. Like most
document-grounded benchmarks, Probity shows the model a window cut from a longer filing, while a
human who read the whole document wrote the label. Nothing in that construction guarantees the
evidence sits inside the window the model receives. When it does not, the item asks for something
its input does not contain, and we score it as though it were merely hard. The audit is this
paper's contribution and the corpus its demonstration, because the same construction assembles
most document benchmarks.

Whether the missing evidence also destabilises those items is a separate question. We
pre-registered a prediction: re-cut each window so the evidence falls inside, and instability
should drop below a threshold we fixed beforehand. Instability did not drop
(Section~\ref{sec:results}): the obvious repair, better windows, did not make the items stable.
The audit stands anyway: an item whose evidence sits outside its window is defective by
construction, however stably a model answers it.

This paper contributes four things. (1) A provenance audit for document benchmarks: it resolves
each label's source by verified content, reports what it could not check, and separates
evidence-absent items from ones that need computation. (2) The measured cost of the defect in our
own corpus, on instability and on apparent cross-model agreement. (3) Probity itself, with a
human-separated oracle protocol and a metric anyone can reimplement. (4) An eight-check list a
benchmark author can run before release, and a measured bound on how little of an accuracy-built
corpus can show instability.

\section{Related work}
\label{sec:related}

\paragraph{Instability as an evaluation axis.} That a model's answer moves between identical calls
is established and measured. \citet{atil-etal-2025-non} find accuracy varying by up to 15\% between
runs across five hosted models at temperature 0 with fixed seeds; \citet{song-etal-2025-good} argue
non-determinism belongs inside an evaluation protocol rather than beside it; and
\citet{potamitis-etal-2025-reasonbench} show that a single observed score can misrank systems. All
three treat the corpus as given and measure how the model moves on it. We ask the question that
comes before those, about the corpus rather than the model: was the item answerable from its input
at all?

\paragraph{Where instability concentrates.} It does not spread evenly.
\citet{yagubyan-2026-coin} report flip rates that vary by task category, on tests they call
underpowered (too few questions per category), so an aggregate figure is misleading rather than
imprecise. In contract review, a nearby domain, \citet{liu-etal-2025-contracteval} break scores
down to the clause level, but for \emph{correctness}: one score per item, and a single score
cannot show whether an answer changes. We break wobble down at the same granularity, and we ask
each item often enough for change to show.

\paragraph{The evaluation as an instrument.} Reporting an evaluation's reliability the way
psychometrics reports a test's is precedented: \citet{cacioli-2026-beyond} and
\citet{contreras-2026-llm} both apply split-half correlation (one half of the runs against the
other) with the Spearman-Brown correction, and our reliability analysis follows their procedure as
the control it is, rather than as a contribution. \citet{cacioli-2026-beyond} is also the nearest
prior result, and it measures a different quantity: the \emph{change} in an item's correctness
between two versions of a model family, for which it reports near-zero item-level agreement across
families, $r = .11$. We measure the \emph{instability} of an item within one fixed model. The two
are different constructs, so there is no tension in one transferring across models while the other
does not.

\paragraph{Financial and legal document benchmarks.} We are not aware of a benchmark evaluating
language models on venture-financing instruments. The nearest by name, VCBench
\citep{chen-etal-2025-vcbench}, predicts founder outcomes from profiles rather than reading
documents; the nearest by document type, MAUD \citep{wang-etal-2023-maud}, annotates merger
agreements and scores mean AUPR over three runs. The broad financial and legal suites, FinBen
\citep{xie-etal-2024-finben} and LegalBench \citep{guha-etal-2023-legalbench}, cover neither
private financing instruments nor stability, and none of the three reports a per-item spread, which
is what would show whether an answer survives being asked twice.

\paragraph{Disaggregated and worst-slice reporting.} Our reporting rule is a refinement of standard
practice, not a discovery. Reporting performance broken down by slice rather than as one number is
established \citep{mitchell-etal-2019-model,liang-etal-2023-holistic}, and reporting against the
worst slice has a long precedent \citep{sagawa-etal-2020-distributionally}. What this benchmark
adds is the two qualifiers that decide whether a worst-slice number means anything: the interval
on the slice, and the count of tasks behind it. Applying them withdraws a headline we would
otherwise have reported (Section~\ref{sec:results}).

\section{The benchmark}
\label{sec:benchmark}

\input{tables/table1_lineup}

\begin{figure}[t]
\centering
\includegraphics[width=\columnwidth]{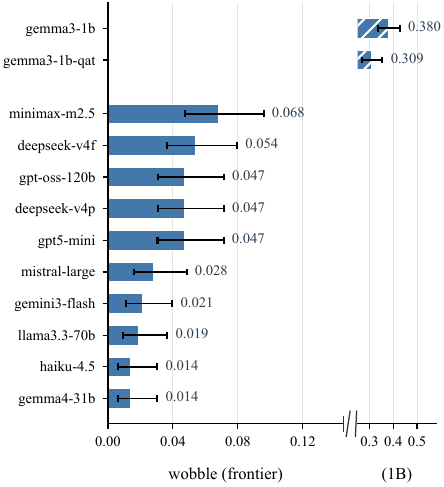}
\caption{Wobble by configuration on the audited corpus, ascending, with Wilson 95\% intervals
($n = 427$ items). The axis is broken: the ten frontier configurations occupy $0$ to $0.145$ and
the two 1B ones resume at $0.25$. A logarithmic axis would fit them on one scale and compress the
order-of-magnitude difference that is the panel's point. The break is drawn instead, and never
where a frontier interval would cross it. Hatching marks the 1B group so the grouping survives a
greyscale print. Item counts and accuracy are in Table~\ref{tab:lineup}.}
\label{fig:lineup}
\end{figure}

First the corpus and how it was labelled, then the audit that checks it, then the metric it is
scored with.

\paragraph{Corpus.} The documents are certificates of incorporation, financing term sheets,
convertible instruments and Form D filings, all from public regulatory sources. From them we build
60 tasks in 8 categories; the largest category holds 16 tasks on priced-equity terms and the
smallest 3 on exit-waterfall computations (who is paid what when the company is sold), with the
full breakdown in Appendix~\ref{sec:datastatement} and the post-exclusion counts in
Table~\ref{tab:categories}. Each task carries between 1 and 18 items, 470 in total.
An item is one document paired with one question about one provision.

\paragraph{Oracle protocol.} A human reads each document and records the answer. The question text
and the answer are stored separately; the model is never shown the labelled value alongside the
question it must answer. Every label carries a validating quote and a pointer to the source filing.
When the supplied text cannot settle an answer, we exclude the item instead of guessing.
The reason for keeping a model out of the labelling loop is specific: an answer key written by a
model can be wrong in exactly the ways the model under test is wrong, and the resulting agreement
would look like accuracy. We do use the configurations in one narrow way, different from writing
the key: when every configuration agrees on an answer the oracle contradicts, the item is listed
as a candidate label error for the author to check against the source document. Nothing a model
proposes enters the corpus until a human has read the filing and upheld it.

\paragraph{What the flag review settled.} It raised 7 items across 5 tasks, 4 with every
configuration agreeing and 3 with a single dissenter. The author adjudicated all 7 as
\textbf{excluded}: none was upheld as an oracle error, none rejected as a vindication of the
oracle; in each case the task as written admitted both readings. No label was changed. The 7 items
are dropped from every reported population, which is why the headline counts are 427 items over 52
tasks rather than 434. Requiring in addition that every configuration hold its answer on at least
18 of its 20 runs admits none of the 7: the weakest configuration's support ranges from 5 to 17
runs; each agreement is between configurations that barely preferred the answer they agreed on.
Table~\ref{tab:flags} reports the floor as a column and does not use it as a filter. Wobble never
references the oracle: no labelling decision can move it. Only the item set changes, and
accuracy is recomputed on the reduced set wherever it appears.

\paragraph{Provenance audit.} We check that each label's validating quote appears in its source
filing and inside the windowed provision the model receives. Whether that check means anything
turns on two construction choices, and each corrects a defect we first shipped and report in
Limitations. Sources are resolved by \emph{verified content}, matching what the document says,
never by identifier convention. Quotes are matched as \emph{spans}, because a validating quote is
sometimes a composite joined by an ellipsis or split by a bracketed insertion. We report coverage
in \emph{three} states: verified present (416), verified absent (5), and not checkable
(49), the state a two-state audit cannot report. 36 of 470 items fail: 31 where the quote
falls outside the model's window only, 2 where it is absent from the source text only, 3 on both
counts.
Pooled over twelve configurations, wobble on flagged items is 0.255, 95\% CI [0.216, 0.298],
against 0.087, [0.080, 0.095] on the \emph{clean} items, those the audit does not flag, a
difference of 0.167, [0.128, 0.211] by Newcombe's method, an interval for the difference of two
proportions. Those intervals treat the pooled pairs as independent and they are not: the flagged
pairs are 36 items crossed with twelve configurations. A bootstrap over \emph{items}, 2{,}000
draws, widens the flagged interval to [0.185, 0.329] and the difference to [0.097, 0.243], still
excluding zero; Table~\ref{tab:stratified} reports it beside the headline and failure-type intervals.
We report headline numbers on the 427 analysed items; Table~\ref{tab:audit} carries
the all-item figures.

\begin{figure}[t]
\centering
\includegraphics[width=\columnwidth]{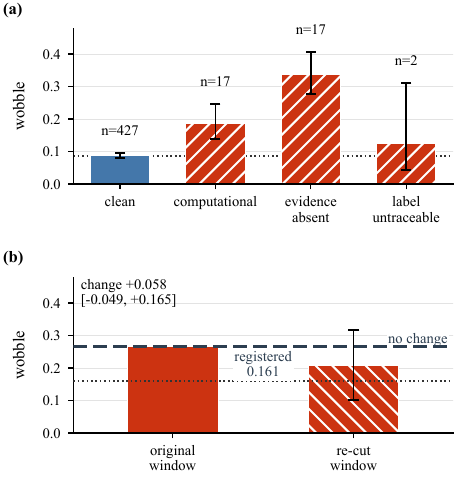}
\caption{The flagged-against-clean contrast and the pre-registered repair. Both panels are on the wobble
scale. (a) Wobble by provenance group with Wilson 95\% intervals, item counts on each bar, ordered
by what the window supplies rather than by effect size; the dotted rule is the clean rate
($n = 427$). Item-bootstrap intervals for the same groups are in Table~\ref{tab:stratified}. The pooled flagged-against-clean contrast is 0.167, [0.128, 0.211] by Newcombe's
method. (b) The pre-registered repair on the 10 repairable items, 20 runs each under twelve
configurations: original window against a re-cut window of equal length. Both bars carry the
flagged colour because re-cutting does not make an item clean. The interval on the re-cut bar is
the \emph{change}, [$-$0.049, 0.165], drawn against the heavy rule at the original rate, the
no-change position, which it crosses. A positive change is a fall in wobble. The
dotted rule is the threshold registered before the runs.}
\label{fig:contamination}
\end{figure}

\paragraph{Two failure types under one flag.} ``The quote is not in the window'' has two causes, and
pooling them makes arithmetic look like a provenance defect. For 17 items the evidence is
genuinely \textbf{absent}: wobble 0.338, [0.277, 0.406]. For 17 others, every number the answer
needs \emph{is} in the window and the stored quote is the annotator's own calculation, a value for
which no verbatim span could exist; all 17 are ownership percentages whose window supplies both
numerator and denominator. These \textbf{computational} items wobble at 0.186, [0.139, 0.245]. Both exceed
the clean rate, by 0.251, [0.189, 0.319] and 0.099, [0.051, 0.158], and the two intervals do not
overlap; the two failure types are separable. Only the first is repairable by re-windowing. A
third group of 2 items has a label we could not trace to any document we hold, though the window
itself is self-sufficient. It sits at 0.125, [0.043, 0.310], a difference of 0.038, [$-$0.044,
0.223], containing zero: where only the \emph{label} is untraceable and the \emph{window} still
holds the answer, the elevation seen in the other two groups does not appear.

Of the 17 evidence-absent items, 10 have their quote verified in a document we hold, and 3 carry a
quote absent from the document we hold, which no widening can surface; those 3 are reported as
oracle-provenance failures rather than windowing failures. We fixed the threshold for a
re-windowing experiment, and the prediction it would test, in writing beforehand. The
pre-registered count of repairable items was 14; Appendix~\ref{sec:deviation} corrects it to 10.
The 4 items that correction removes from the repair stay in the evidence-absent group wherever its
wobble is reported; the correction governs repair eligibility alone. Without those 4, the
remaining 13 wobble at 0.237, [0.177, 0.310] (Wilson), above the clean rate by 0.150, [0.089, 0.223] by Newcombe.

\paragraph{What the flagged items have in common.} Flagged items are harder: majority accuracy on
them is 0.722 against 0.909 on the rest, a difference of 0.187, [0.146, 0.232]. They are also
overwhelmingly numeric, concentrated in two computational categories;
Appendix~\ref{sec:composition} gives the breakdown. The exclusion removes 8 tasks outright, and 7
of those 8 are cap-table or waterfall computations. A corpus that dropped them would lose most
of its arithmetic.

\input{tables/table11_stratified}

\paragraph{The contrast, stratified.} Because flagged items are harder and mostly numeric, the
pooled contrast could in principle be an answer-type or difficulty effect rather than a provenance
effect. Table~\ref{tab:stratified} takes both objections on their own terms. Inside the numeric
stratum, which holds 83.3\% of the flags, flagged items wobble at 0.283 against 0.093 on clean
numeric items, a difference of 0.190, [0.143, 0.241] by Newcombe; with the two 1B
configurations dropped, the difference is 0.121, [0.081, 0.168]. The item-bootstrap intervals widen
every one of these and the numeric and frontier-only differences still exclude zero. The
non-numeric flagged group has 6 items, and its difference contains zero; we report it for
completeness. A category-matched comparison is not constructible on this corpus: the audit flags
25 of 33 cap-table items and 6 of 8 exit-waterfall items, so the flagged items largely \emph{are}
those categories. Binary wobble says whether an item flips at all; the dispersion rows of
Table~\ref{tab:stratified} say how much, and the flagged-against-clean gap holds on that measure
too, 0.104 against 0.019 pooled.

\paragraph{A property that predicts instability without the flag.} A simpler, mechanical question
separates the corpus without using the validating quote or the span check that defines the flag:
is the value the oracle records stated anywhere in the window? Where the window states the answer,
wobble is 0.085, [0.071, 0.102]; where it does not, 0.264, [0.225, 0.306]. The audit did \emph{not}
flag 8 of the items whose window fails to state the answer, and they still wobble at 0.167,
[0.105, 0.254]; the property carries signal the flag misses. Appendix~\ref{sec:composition}
gives the construction and its robustness check.

\paragraph{Wobble.} For an item answered $n$ times by one model, let $c$ be the number of responses
whose normalised answer equals the most frequent one. The item is unstable when $c < n$, and wobble
for a set of items is the share that are unstable, weighted by item and not by task. Normalisation
is type-aware: numeric answers compare as numbers, enumerated answers against a fixed vocabulary,
dates as dates. Correctness is scored separately, as majority-vote agreement with the oracle over
the same $n$ responses, and is never averaged with wobble. We use $n = 20$ at temperature 0.7, fixed
once for all twelve configurations. The value is deliberate: non-zero because sampling is the
regime wobble measures, below the default of 1.0 the directly served APIs apply when a request
sets none, and above the near-greedy settings practitioners deploy when they want consistency. An
item can therefore be unstable while mostly correct, and stable while always wrong.

Concretely: twenty runs on one item return \$1{,}250{,}000 seventeen times, 1250000 twice and
\$1{,}500{,}000 once. Normalisation strips currency marks and separators, so the first two forms
are one number and $c = 19$ of $n = 20$: the item is unstable. Its modal answer matches the
oracle, so the same item counts as correct.

\paragraph{Models.} Twelve configurations, listed in Table~\ref{tab:lineup}, spanning eleven
distinct models over three serving paths: two run locally, three through a first-party API, and
seven through a commercial routing layer. The routing layer adds non-determinism of its own, and
our design cannot tell it apart from the model's own sampling.

\section{What the instrument can resolve}
\label{sec:instrument}

Two properties of the instrument come first: whether the measurement repeats, and where it can
show anything at all.

\begin{figure}[t]
\centering
\includegraphics[width=\columnwidth]{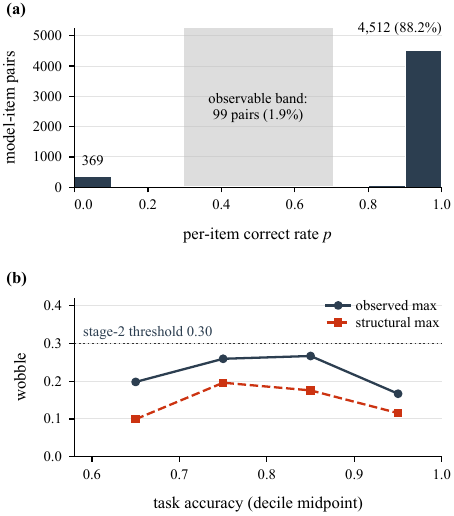}
\caption{(a) Distribution of the per-item correct rate $p$ over the 5{,}116 measured pairs of one
configuration with one audited item. The shaded region is the observable band,
$0.3 \le p \le 0.7$. The labelled bar is the bin $p \geq 0.9$; the 86.0\% quoted in
Section~\ref{sec:instrument} is the narrower set with $p = 1$. (b) Largest observed and largest
structurally reachable wobble per accuracy decile, the structural maximum being the most wobble an
item pool with the observed $p$ values could produce. The axis starts at 0.6 because no task is
answered correctly less often than that, and the four deciles shown are every one containing a
task.}
\label{fig:observability}
\end{figure}

\paragraph{Reliability.} Split-half correlation with the Spearman-Brown correction
\citep{cacioli-2026-beyond,contreras-2026-llm} gives a median of 0.903 across configurations;
Appendix~\ref{sec:appendix} gives the procedure and the per-configuration values. This is a control
on the measurement rather than a result about models: the ranking of tasks by instability is
reproducible from half the data, and a small difference between two tasks is not automatically
noise.

\paragraph{Observability.} Wobble can show only where the model is uncertain, and this corpus
gives it little room. 91.3\% of model-item pairs are answered identically on all 20 runs and
86.0\% are answered correctly on all 20; an identical answer is usually a correct one. A further
6.3\% are never correct, though not always stably: an item can be wrong a different way each run.
Only 1.9\%, 99 of 5{,}116 pairs, sit in the band $0.3 \le p \le 0.7$, $p$ being the item's correct
rate over its 20 runs, where an item can flip at all (Figure~\ref{fig:observability}a); the
half-open reading, $0.3 \le p < 0.7$, gives 85 and 1.7\%, and we report both because $p$ takes
only the values $k/n$ and the boundaries are populated. A corpus assembled to test accuracy is
therefore largely silent about stability. Figure~\ref{fig:observability}b makes the ceiling
explicit: among tasks answered correctly at least 90\% of the time, the most wobble the observed
items could ever show is 0.115, below the 0.30 threshold that would define a correct-but-unstable
task. That region was unreachable before any model ran.

\section{What the benchmark shows}
\label{sec:results}

\input{tables/table2_transfer}
\input{tables/table4_claims}

\paragraph{Instability is partly shared across models.} Take a frontier configuration's five
least-stable \emph{items} and ask how many fall inside another configuration's ten least-stable;
call the share that do \emph{transfer}. We use items rather than tasks because most tasks never
flip: a task-level ranking cannot fill a top five on 42 of the 90 ordered pairs, against 36 at
item level. Ranking also needs a graded score, which the binary flip flag is not; we rank by
the share of runs disagreeing with the most frequent answer. Table~\ref{tab:tasklevel} runs both
units. The task-level reading finds the same effect on a coarser unit whose overlap expected by
chance is 0.192 rather than 0.023: its ratio is 2.3 where the item-level one is 8.9, at the
same permutation $p$.

\begin{figure}[t]
\centering
\includegraphics[width=\columnwidth]{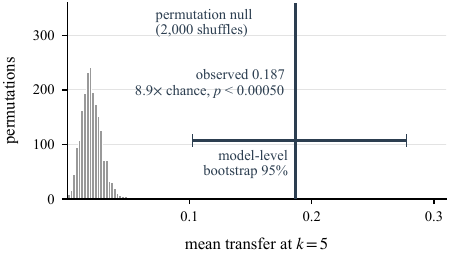}
\caption{Mean transfer at $k = 5$ against its permutation null, on 425 clean items. The histogram
is the null itself, 2{,}000 permutations that shuffle each configuration's dispersion values among
its own item positions, preserving its value multiset and therefore its effective $k$. The observed
0.187 sits outside every draw; the reported $p < 0.0005$ is the resolution of 2{,}000
permutations rather than a computed tail. The horizontal span is the 95\% bootstrap interval,
[0.102, 0.278], resampled over the ten \emph{configurations} rather than the 90 ordered pairs,
which are ten models crossed with themselves and are not independent.}
\label{fig:transfer}
\end{figure}

Over the 90 ordered pairs among the ten frontier configurations and 425 clean items, the mean
transfer is 0.187 with a model-level bootstrap interval of [0.102, 0.278], and the configurations
share 84 items against 9.4 expected by chance, a ratio of 8.9 (Table~\ref{tab:transfer}). The item
set is 425 rather than 427 because an item enters only where every configuration measured it, and
2 items have a configuration whose 20 responses all failed to parse. A
permutation null that shuffles each configuration's values among its own positions puts both
statistics at $p < 0.0005$ over 2{,}000 permutations, the resolution of that many draws rather
than a computed tail (Figure~\ref{fig:transfer}). The bootstrap
resamples the ten \emph{configurations}, not the 90 pairs, which are not independent: they are ten
models crossed with themselves.

The median is the natural summary and the wrong one here. Overlap of a five-item set can only take
multiples of $0.2$, and most pairs share nothing; the median reads 0.000 at $k = 3$ and
$k = 4$. Its own permutation null is 0.000 at every percentile, separating only zero from
non-zero, and its bootstrap interval at $k = 5$ contains zero. The mean and the hits ratio are
stable across $k$ (10.2, 9.0, 8.9): what depends on $k$ is the statistic, not the effect.
Agreement is also not a majority phenomenon: 26 of 90 pairs share any item at $k = 3$, 51 of 90 at
$k = 5$.

\paragraph{Four controls.} The standing objection is that models agree on which items are
\emph{hard} rather than on which are unstable. Holding difficulty fixed answers it: within the
accuracy stratum (items of similar difficulty) that contains 407 of the 425 items, transfer is 5.6
times chance: part of the raw effect is difficulty and most is not. Rank-biased overlap
\citep{webber-etal-2010-similarity} is defined for rankings of unequal length and is no
artefact of where a top-$k$ was cut. It gives a median 0.063 at $p = 0.9$: stricter, and
informatively so, because the configurations agree substantially on \emph{which} items are
unstable and weakly on their \emph{order}. Dropping each configuration in turn keeps the ratio
between 7.8 and 10.2 (Table~\ref{tab:loo}). Splitting by serving path is the one control that
finds something: the three directly served configurations agree at a mean of 0.567 against 0.152
among routed pairs, and 0.550 once the two same-family configurations are excluded. Six ordered
pairs from three configurations is underpowered, but it suggests part of the shared instability is
infrastructure rather than models.

Recomputing the same statistic before the audit exclusion gives a mean of 0.238 and a ratio of 11.0
at $k = 5$, against 0.187 and 8.9 after: excluding the flagged items removes a fifth of the
apparent sharing: contamination inflates cross-model agreement without being its main cause.

\paragraph{The repair fails its prediction.} We registered a threshold before running it:
repaired wobble would fall below 0.161, the midpoint of the clean and flagged rates. Each of the 10 repairable items was re-cut to a window of the same character length, centred
on its validating quote and verified by the audit's own matcher to contain it, then answered 20
times by all twelve configurations. Equal length matters, because a wider window would confound the
repair with extra context.

The repair achieved what it was built to achieve. Of the 8
repairable items whose answer is a number, none had that answer stated in its original window; all
8 do in the re-cut window at the same length. The configurations were shown the
answer and went on disagreeing with themselves about it. What follows is a successful
intervention with no measurable effect.

The repaired wobble is 0.208, [0.145, 0.289], against 0.267, [0.196, 0.352] on the same items in
their original windows: a change of 0.058, [$-$0.049, 0.165], containing zero, and above the
registered threshold (Figure~\ref{fig:contamination}b, Table~\ref{tab:repair}). The prediction
fails. Restricting to the 107 of 120 cells where both conditions have all 20 responses moves it to
0.187, still above 0.161; the result is not an artefact of the original runs' missing
responses. We therefore withdraw the causal claim and report the association as correlational
only. The pre-registration commits us to this reading, with no reframing, re-slicing or
threshold-moving after the fact. One task of the three does move a long way, option
pool shuffle falling from 0.611 to 0.306 while the other two are flat or slightly worse; that
split was not pre-registered and we do not build on it. The items are unstable both before and
after the repair: whatever destabilises them survives having the answer put in front of the
model.

\paragraph{The worst-category claim does not survive.} A natural recommendation is that a model's
headline wobble understates its worst clause category by some multiple, and the point estimates
support it: the median worst-to-mean ratio is 2.64. It does not survive a conservative reading:
taking the largest Wilson lower bound (the interval's cautious end) over estimable categories
against the Wilson upper bound of the mean avoids picking a category on the same data used to
score it, and under that rule no configuration's worst category clears the mean, whether we
require three tasks or five before a category is estimable. The category with the largest
apparent effect is also the smallest, a property of this corpus rather than of the models.
Table~\ref{tab:claims} lists it beside every claim we tested against this benchmark and could not
support, our registered prediction at the top.

\section{Using the benchmark}
\label{sec:using}

\paragraph{Provenance checklist.} Eight mechanical checks a benchmark author can run before
release, each corresponding to a failure we found in our own corpus.
(1)~Resolve each item's source document by \emph{verified content}, never by filename or identifier
convention, and report the count per route.
(2)~Verify the label's validating span occurs in that document.
(3)~Verify it occurs inside the window the model is shown; a different check, and the one that
predicts instability.
(4)~Report three-way coverage, including \emph{not checkable}: saying what you could not check is
stronger than implying you checked everything.
(5)~Emit a specific reason per rejection; a bare count cannot be audited.
(6)~Confirm the checker passes known-good input before acting on a failure count: a false
rejection discards real data as silently as a false pass invents it.
(7)~Separate items whose evidence is \emph{absent} from those whose answer must be \emph{computed}
from evidence present; the rates and the repairs differ.
(8)~Fix any repair experiment's threshold in writing before collecting the data that decides it.

\paragraph{Reporting rule.} Report per-category wobble with intervals and the task count for each
category beside any aggregate, and never quote a worst-category figure from a category with fewer
than five tasks. Disaggregated and worst-slice reporting has precedent (Section~\ref{sec:related});
the two qualifiers decide whether such a figure means anything, and applying them withdrew one of
ours.

\paragraph{Corpus design rule.} Build items that land in $0.3 \le p \le 0.7$ deliberately, since
that band holds 1.9\% of this corpus's measurements and bounds what any instability study on it
can resolve. Items requiring a computation over values stated in several places are the natural
candidates and the hardest to window; a successor corpus should record a computed answer's
provenance as a set of spans instead of one.

\paragraph{Release.} We release the corpus, the oracle with its validating quotes, all 112{,}800
raw responses collected of which 111{,}800 are analysed, the scoring engine, and a standalone
verifier that reproduces every number here while importing nothing from the benchmark
package. The provenance audit and its checker's selftests run as part of the release: a user
can reproduce the exclusion instead of taking it on trust, and can see the checker demonstrated on
input it must reject and accept. We also release the blind re-annotation pack with
its protocol and adjudication rule. \artifactpointer

\section{Conclusion}

A benchmark assembled from windowed extracts can ask questions its own inputs cannot answer. Our
audit finds those items, separates them from answers computable from evidence present, and
reports what it could not check. We release the corpus, with a
correctness-independent wobble metric and split-half reliability 0.90. Where evidence is genuinely
absent, wobble is 0.338 against 0.087; excluding all flagged items cuts apparent cross-model
agreement by about a fifth.
Re-cutting windows where possible did not lower wobble to the registered level, moving it
0.058, [$-$0.049, 0.165]; we report the association as correlational, the mechanism as untested,
and the audit stands either way. It cannot resolve whether models produce
correct-but-unstable answers: only about 2\% of its measurements sit where instability can show.%
\label{mainend}

\section*{Limitations}

A single annotator produced every oracle label, and no inter-annotator agreement is reported. The
automated provenance audit described in Section~\ref{sec:benchmark} checks that each label's
validating quote exists in the source document and falls inside the model's window; it does not
check that the label is the right answer. The model-flagged review in
Section~\ref{sec:benchmark} raised 7 candidate label errors, and none was upheld. That review can
only surface errors the models happen to disagree with the oracle about; a wrong label that every
configuration reproduces is exactly the one it cannot reach.

Each of the 7 was read against its source text individually, and each was excluded as ambiguous on
the task as written rather than mislabelled. Two windows state two seniority structures, one split
by a date the question does not fix and one with its ranking clause truncated inside the window.
Three carry an explicit participation-cap proviso inside the sentence the label was anchored on,
and the excerpt alone does not settle whether that defined cap binds. One states two preference
multiples across sub-series the question does not distinguish. And one holds prices from two
different issuances without fixing which round is asked about. Every one of these exclusions removes an item the configurations answered \emph{wrongly}, and the
review therefore raises measured accuracy. With none upheld and none rejected, it vindicates neither the oracle nor
the models; it finds items that under-determine their answers.

No second reading of the corpus has been performed, so this paper reports \emph{no} agreement
coefficient of any kind, and any that appears in later work on this corpus should be read as
intra-annotator unless its author says otherwise. That is the one part of this work a reader cannot
check by re-running the artifact, so we release what an independent run needs: the stratified
sample manifest; the blind pack of 154 items, original labels removed, row order randomised under
a recorded seed; the answering protocol; the adjudication rule; and the scorer, which refuses to
report an agreement figure on an unfilled sheet rather than scoring empty against empty. Any reader can therefore run an agreement study against our corpus without our
participation, and can disagree with our labels on the record. We would rather make that possible
than ask to be believed.

The audit itself had defects, found by auditing its rejections rather than its pass rate, and both
are properties this paper's numbers depend on. Its source lookup resolved documents by item
identifier while they are stored under company and accession prefixes, so 150 of 470 items were
never opened and their quote check silently did not run. Its quote matcher tested the raw text of
composite quotes, which contain separators occurring in no source document, so 8 items were flagged
for the matcher's construction. Both are repaired here and both moved published numbers. We report
them because a reader has no way to know from a corrected artifact how much of it was ever wrong.

The re-windowing repair covers only the 10 repairable items of the 17 whose evidence is absent, not
all 36 flagged items, and every statement about it names that subset. Three of the other 7 carry a
quote absent from a document we hold and no widening can surface it; 4 are computational under the
pre-registration's definition and were excluded from the repair while staying in the
evidence-absent figures, for the reason given in Appendix~\ref{sec:deviation}. Ten items is a small experiment and it is the whole of the population
the pre-registration made eligible, so the negative result is reported with that width rather than
as a general finding about windowing. A repair that failed on 10 items does not establish that
re-windowing never helps.

The original runs the repair is measured against were collected before the empty-response refill
described in the artifact. As a result, 12 of the 120 paired cells have fewer than 20 responses in
the original condition, one as few as 6, and 13 are short in one condition or the other. A smaller
denominator makes instability harder to observe, which biases the original condition toward
looking more stable and the repair effect toward looking smaller. Both readings are reported in Section~\ref{sec:results} and neither crosses
the registered threshold.

All results are at a single temperature, 0.7. Temperature effects are out of scope, and instability
at temperature 0 is a different measurement with different causes \citep{atil-etal-2025-non}.

Seven of the twelve configurations are served through a commercial routing layer. That layer
introduces non-determinism of its own, from batching and from backend substitution, and this design
does not separate it from the model's sampling. Two configurations are quantizations of one base
model, so the twelve
configurations span eleven distinct models and the two are not independent observations.

Our run manifests pin the corpus by content hash, the run count and the temperature, but not the
model build. For the ten hosted configurations the provider and model identifier are recorded and a
silent substitution behind them is a known and stated risk. For the two locally served
configurations there is no equivalent: the local tag is mutable and the install can be removed, and
ours had been by the time the repair condition was run. We re-obtained them and tested behavioural
equivalence: re-run the original windows, then compare each item's most frequent answer against a
resampling null built from the stored responses. The null is the right comparison because a
configuration unstable on a third of items changes its most frequent answer between two samples
without anything having changed. Both configurations are consistent with the published arm on that
test. It establishes behavioural equivalence on the repaired items, not identity of weights, and we
report it as the weaker claim it is. A manifest that versions the data but not the model cannot
support a reproduction claim for a locally served model, and ours should have recorded the local
digest.

Category task counts are unbalanced, from 3 to 16 before the provenance exclusion and from 1 to 16
after it, and the category with the largest apparent effect is among the smallest. Two categories
keep fewer than three tasks on the audited corpus, too few to estimate, and no claim in this paper
rests on them.

Wobble and accuracy are computed from the same 20 responses per item and are structurally related:
an item answered correctly on every run cannot be unstable. Their correlation is therefore not
evidence about models, and we report it only to concede the point.

The corpus is English-only and drawn from United States jurisdiction filings. Terms, drafting
conventions and the meaning of individual provisions differ elsewhere.

470 items over 60 tasks, reduced to 427 over 52 by the audit and the adjudication, is a small
corpus. The interval widths in Table~\ref{tab:categories} reflect that honestly. Most pairs of
category cells have overlapping intervals, which is why this paper asserts no ordering between
categories. The serving-path split rests on three directly served configurations and six ordered
pairs,
which is the narrowest result in the paper and is reported as underpowered wherever it appears.

The audit has been run on one corpus. The released checker takes each item's window, validating
quote and source text, and running it against a second windowed benchmark is the natural next
test of whether the failure classes found here generalise.

\section*{Acknowledgments}

I thank the three FinNLP reviewers for comments that improved the paper.

\section*{Use of generative AI}

Generative AI assistance was used for analysis code and for drafting this paper. The author
verified every result reported here against the released artifact and takes full responsibility for
the content.

\bibliography{anthology,custom}

\appendix

\input{sections/data_statement}

\section{What the flagged items have in common}
\label{sec:composition}

The flagged items are overwhelmingly numeric, carrying 83.3\% of the flags while numeric-answer
items are 32.0\% of the corpus, a flag rate of 0.203 against 0.009 on items with a binary answer.
They concentrate in two clause categories: 25 of 33 cap-table items and 6 of 8 exit-waterfall items
are flagged, while four of the eight categories contain no flagged item at all.

The answer-stated-in-window property of Section~\ref{sec:benchmark} is computed from the answer and
the window alone and is decidable for 137 of the items. Reading numerals written as words is what
makes it decidable on a filing at all; dropping that pass moves the two rates from 0.085 and 0.264
to 0.079 and 0.252, so the separation holds either way.

Reliability is
computed by splitting each item's 20 runs into odd and even halves, recomputing the full
task-by-model matrix on each half, correlating a configuration's two half-vectors across tasks and
applying the Spearman-Brown correction. Figure~\ref{fig:reliability} gives the per-configuration
values behind the median quoted in Section~\ref{sec:instrument}.

\section{A correction to the repairable count, against our own interest}
\label{sec:deviation}

The pre-registration records 14 repairable items, because retrieving the cited filing resolved 4
further ones. Those 4 are computational by the pre-registration's own definition, and were
classified as evidence-absent only because the classifier reads operands out of the stored quote
and theirs is a pointer to a table naming none. Their windows carry a complete labelled
derivation, and the arithmetic checks against the oracle to two decimals, so re-cutting the window
around the quote would replace a self-sufficient derivation with source text that does not contain
the operands. The pre-registration excludes computational items from the repair in advance and in
those words, so the corrected target is 10, which is exactly the registered floor of 10 and clears
it without margin. Runs had already been collected on the 4 when the misclassification was found,
and were discarded rather than reported. That is a deviation, and we record it here rather than in
the pre-registration, which is left as written.

\section{Full results}
\label{sec:appendix}

This appendix holds the disaggregated results the main text summarises, and the record of what the
provenance exclusion changed.

Table~\ref{tab:categories} gives wobble for every configuration in every clause category, with a
Wilson interval and a task count for each cell. It is the reporting rule of
Section~\ref{sec:using}, applied cell by cell. Two of the eight categories retain fewer than three tasks after the
exclusion. They are printed rather than dropped, because their size is the reason no claim rests
on them and a reader should be able to see it.

\input{tables/table3_categories}

Table~\ref{tab:audit} places every headline quantity under both readings side by side. The effect
of excluding the 36 flagged items can be read off it directly, including on the quantities it
barely moved.

\input{tables/table5_audit}

Table~\ref{tab:repair} gives the re-windowing repair by task, behind the pooled result in
Section~\ref{sec:results} and Figure~\ref{fig:contamination}b.

\input{tables/table6_repair}

Table~\ref{tab:tasklevel} runs the transfer statistic under both ranking units, which is the
comparison behind ranking items rather than tasks. Tables~\ref{tab:serving} and~\ref{tab:loo} give
the two splits that test whether the effect is carried by part of the lineup.

\input{tables/table7_tasklevel}

\input{tables/table8_serving}

\input{tables/table9_loo}

Table~\ref{tab:flags} is the model-flag worklist described in Section~\ref{sec:benchmark}: every
item where the configurations agree with each other and contradict the oracle. It is printed with
its adjudication column so a reader can see both what was raised and what was settled.

\input{tables/table10_flags}

\begin{figure}[t]
\centering
\includegraphics[width=\columnwidth]{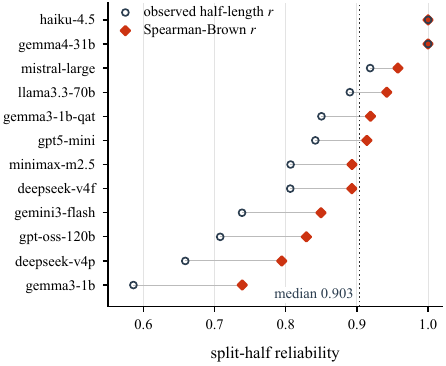}
\caption{Split-half reliability by configuration on the audited corpus, over the 49 tasks with
non-zero wobble in at least one configuration. Circles are the observed correlation between the
odd-run and even-run task vectors; diamonds are the Spearman-Brown correction to full length. The
dotted rule marks the median corrected value. Two configurations sit at exactly 1.000 on both
measures, so their circle and diamond coincide. They have four and five tasks with non-zero wobble
out of the 49, so both half-vectors are almost entirely ties and the rank correlation is
correspondingly easy to saturate; the reported statistic is the median, which does not depend on
them.}
\label{fig:reliability}
\end{figure}

Figure~\ref{fig:reliability} gives the per-configuration split-half reliability behind the median
quoted in Section~\ref{sec:instrument}.

\end{document}

%% file: tables/table1_lineup.tex
\begin{table}[t]
\centering
\scriptsize
\setlength{\tabcolsep}{3pt}
\begin{tabular}{l l S[table-format=3.0] S[table-format=3.0] S[table-format=1.3]}
\toprule
Model & Serving & {Items} & {Unstable} & {Acc.} \\
\midrule
gemma4-31b & routed & 426 & 6 & 0.965 \\
haiku-4.5 & direct & 425 & 6 & 0.962 \\
llama3.3-70b & routed & 426 & 8 & 0.953 \\
gemini3-flash & routed & 426 & 9 & 0.965 \\
mistral-large & routed & 427 & 12 & 0.944 \\
gpt5-mini & routed & 427 & 20 & 0.965 \\
deepseek-v4p & direct & 426 & 20 & 0.969 \\
gpt-oss-120b & routed & 426 & 20 & 0.965 \\
deepseek-v4f & direct & 427 & 23 & 0.967 \\
minimax-m2.5 & routed & 427 & 29 & 0.963 \\
\midrule
gemma3-1b-qat & local & 427 & 132 & 0.663 \\
gemma3-1b & local & 426 & 162 & 0.631 \\
\bottomrule
\end{tabular}
\caption{Twelve configurations on the 427 items that pass the provenance audit (52 of 60 tasks retain at least one item), 20 runs per item at temperature 0.7. \emph{Unstable} counts items whose answer is not identical across all 20 runs; accuracy is majority-vote correctness over the same runs. Wobble, the ratio of the two counts, is Figure~\ref{fig:lineup}. The rule marks the order-of-magnitude gap; the two 1B rows are quantizations of one base model, so the twelve configurations span eleven distinct models.}
\label{tab:lineup}
\end{table}

%% file: tables/table11_stratified.tex
\begin{table*}[t]
\centering
\footnotesize
\setlength{\tabcolsep}{2.5pt}
\begin{tabular}{l l r r r r r r}
\toprule
 & Stratum & Cfg. & f/c & Flagged & Clean & $\Delta$ (Newc.) & $\Delta$ (boot) \\
\midrule
wobble & pooled & 12 & 36/427 & 0.255 [.185,.329] & 0.087 [.077,.098] & 0.167 [.128,.211] & [.097,.243] \\
wobble & numeric & 12 & 30/118 & 0.283 [.205,.378] & 0.093 [.073,.115] & 0.190 [.143,.241] & [.109,.282] \\
wobble & non-numeric & 12 & 6/309 & 0.111 [.042,.181] & 0.085 [.074,.097] & 0.026 [$-$.029,.119] & [$-$.046,.103] \\
wobble & pooled & 10 & 36/427 & 0.144 [.072,.231] & 0.036 [.027,.045] & 0.109 [.075,.149] & [.036,.198] \\
wobble & numeric & 10 & 30/118 & 0.163 [.070,.270] & 0.043 [.026,.063] & 0.121 [.081,.168] & [.026,.229] \\
wobble & non-numeric & 10 & 6/309 & 0.050 [.017,.083] & 0.033 [.023,.044] & 0.017 [$-$.017,.104] & [$-$.022,.056] \\
\midrule
wobble & evidence-absent & 12 & 17/427 & 0.338 [.201,.495] & 0.087 [.077,.098] & 0.251 [.189,.319] & [.115,.409] \\
wobble & computational & 12 & 17/427 & 0.186 [.157,.216] & 0.087 [.077,.098] & 0.099 [.051,.158] & [.068,.130] \\
wobble & label-untraceable & 12 & 2/427 & 0.125 [.083,.167] & 0.087 [.077,.098] & 0.038 [$-$.044,.223] & [$-$.010,.087] \\
\midrule
disp. & pooled & 12 & 36/427 & 0.104 [.078,.130] & 0.019 [.016,.022] & {--} & 0.085 [.059,.112] \\
disp. & pooled & 10 & 36/427 & 0.041 [.017,.070] & 0.007 [.005,.009] & {--} & 0.034 [.010,.062] \\
\bottomrule
\end{tabular}
\caption{The flagged-against-clean contrast, stratified. Brackets on the Flagged and Clean cells are 95\% intervals from a bootstrap over \emph{items} (2{,}000 draws, resampling flagged and clean items independently; narrower than Wilson where items are homogeneous, degenerate at $n = 2$), the resampling unit because each item contributes one pair per configuration and those pairs are not independent; the difference column carries the Newcombe interval on the two Wilson intervals beside the item-bootstrap interval for the same difference. \emph{Cfg.} 12 is every configuration, 10 drops the two 1B configurations. The three failure-type rows split the flagged items by cause (Section~\ref{sec:benchmark}), each against the whole clean set, pooled configurations only. Newcombe's method applies to proportions; dispersion is a mean, so its difference carries the bootstrap interval only. \emph{Dispersion} is the share of runs disagreeing with the modal answer, averaged over the same model-item pairs as wobble. The numeric stratum holds 83.3\% of the flags; the non-numeric flagged group has 6 items and the label-untraceable group 2, reported for completeness, not weight.}
\label{tab:stratified}
\end{table*}

%% file: tables/table2_transfer.tex
\begin{table}[t]
\centering
\scriptsize
\setlength{\tabcolsep}{2pt}
\begin{tabular}{S[table-format=1.0] c S[table-format=1.3] S[table-format=2.2] r c r}
\toprule
{$k$} & {Mean} & {Chance} & {Ratio} & $p$ & 95\% CI & Trunc. \\
\midrule
3 & 0.144 & 0.014 & 10.23 & $<$ 0.00050 & [.042,.261] & 0/90 \\
4 & 0.161 & 0.019 & 9.01 & $<$ 0.00050 & [.066,.266] & 18/90 \\
5 & \textbf{0.187} & 0.024 & 8.91 & $<$ 0.00050 & [.102,.278] & 36/90 \\
\bottomrule
\end{tabular}
\caption{Cross-model transfer of instability, frontier configurations only (10 models, 90 ordered pairs, 425 clean items), at each of the 3 values of $k$ the corpus supports. Each pair asks what share of configuration $A$'s $k$ least-stable \emph{items} fall inside $B$'s $2k$ least-stable. \emph{Chance} is the expected share under the effective $k$ after truncation, \emph{Ratio} is observed hits over expected, and the permutation $p$ is over 2{,}000 shuffles of each configuration's own values. The interval is a bootstrap over the ten \emph{configurations}, not the 90 pairs, which are not independent. \emph{Trunc.} counts pairs where a configuration had fewer than $k$ unstable items and the set was truncated rather than padded. $k=5$ is the pre-specified value. The claims this benchmark could not support are Table~\ref{tab:claims}.}
\label{tab:transfer}
\end{table}

%% file: tables/table4_claims.tex
\begin{table*}[t]
\centering
\footnotesize
\setlength{\tabcolsep}{4pt}
\begin{tabular}{>{\raggedright\arraybackslash}p{0.27\textwidth} >{\raggedright\arraybackslash}p{0.15\textwidth} >{\raggedright\arraybackslash}p{0.36\textwidth} >{\raggedright\arraybackslash}p{0.15\textwidth}}
\toprule
Claim tested & Point estimate & Under the conservative reading & Status \\
\midrule
Re-windowing lowers wobble to the registered level & 0.208 after, 0.267 before & change 0.058 [$-0.049$, $0.165$] contains zero, and 0.208 exceeds the registered 0.161 & not supported \\
Worst category runs $N\times$ the mean & median 2.64$\times$ & 0.60$\times$ at the Wilson bounds; 0/12 above $2\times$ & not supported \\
The gap grows as models improve & slope 3.43 & intercept [$-0.082$, $0.049$] contains zero ($R^2$ 0.680, $n$ = 10) & not supported \\
Instability is not merely difficulty & $\rho = -0.487$ & residual $\rho = 0.189$ after the structural term & conceded \\
Correct-but-unstable answers do not occur & 0 tasks observed & ceiling 0.115 $<$ 0.30 threshold & unreachable by construction \\
\bottomrule
\end{tabular}
\caption{Claims tested against this benchmark and not supported by it, our own pre-registered prediction first. \emph{Point estimate} is the number a reader of the aggregate would quote; \emph{under the conservative reading} is what survives attacking it. For the worst-category claim that means the largest Wilson lower bound over estimable categories against the Wilson upper bound of the mean, which avoids selecting a category on the same data used to score it. \emph{Conceded} marks a claim we do not make; \emph{unreachable by construction} marks one this corpus cannot test at all, for the reason given in Section~\ref{sec:instrument}.}
\label{tab:claims}
\end{table*}

%% file: sections/data_statement.tex
% paper/sections/data_statement.tex
% The CFP asks dataset papers for a data statement. Full version ships in the release repository as
% DATA_STATEMENT.md; this is the appendix summary. Every number is recounted from the artifact by
% audit/ds_facts.py, not copied from the prose of this paper. Two items are UNRESOLVED and are
% printed as unresolved rather than filled with a plausible value.
\section{Data statement}
\label{sec:datastatement}

\paragraph{Curation and sources.} Probity asks models to read venture-financing documents and answer
questions whose answers carry money. One document class: public filings retrieved from the U.S.\ SEC
EDGAR system, covering 965 distinct filings from 863 distinct filer CIKs. Items are identified as
\texttt{CIK\_accession}, so each states which filing it came from and can be re-fetched from the
EDGAR Archives. Two artifacts are stored per item: the full document as retrieved
(1{,}145 files, 60.5\,MB) and the windowed extract actually shown to the model
(524 files, 0.76\,MB). The distinction between them is this paper's subject.

\paragraph{Composition.} 60 tasks over 470 items in 8 categories: priced equity (16 tasks),
convertibles (12), cap table (7), rights and governance (7), founder equity (5), regulatory (5),
risk flags (5), exit waterfall (3). Twelve configurations spanning eleven distinct base models over
three serving paths (two local, three direct, seven routed), 20 samples per item at temperature 0.7:
112{,}800 responses collected, 111{,}800 analysed after 1{,}000 parse failures.

\paragraph{Exclusions.} Three layers, in order: 470 items as annotated; minus 36 removed by the
provenance audit (31 quote-not-in-window, 3 both, 2 quote-not-in-source); minus 7 removed by author
adjudication; leaving 427. Of 60 tasks, 52 retain at least one item. Audit coverage is reported three
ways rather than as a pass rate: 416 verified-present, 49 not-checkable, 5 verified-absent. The
audit's own two defects (a source lookup that never opened 150 of 470 items, and a matcher that
flagged 8 items for its own construction) are reported in the Limitations section rather than
silently repaired, because a reader cannot tell from a corrected artifact how much of it was wrong.

\paragraph{Annotation.} One annotator produced every label. There is no second reading and this
paper reports no inter- or intra-annotator agreement coefficient. Two consequences are stated exactly
rather than loosely: the 7 items removed at adjudication were removed by the same person who produced
the labels, so that step has no independent arbiter; and if the prepared pack is ever re-read by that
same person, the result is \emph{intra}-annotator agreement and must be reported in those words,
because only an independent reader yields \emph{inter}-annotator agreement. The blind re-annotation pack
(154 items over 19 of 60 tasks, sized so that every category $\times$ answer-type stratum is
represented), its protocol and its adjudication rule are released so a second reader can be run by
anyone; the sitting has not happened. Every label ships with a validating quote and its source
filing, which makes a label traceable to a document. That is not a substitute for a second annotator.

\paragraph{Personal and identifying data.} The corpus is corporate. It names 313 distinct companies,
retained deliberately because they are what makes a label auditable. Some labels do characterise a
named company's terms: the five \texttt{flag\_*} tasks record readings such as an off-market
liquidation preference or an uncapped participation against specific filings, 39 items in all. Each is one
annotator's reading of one clause, carrying a validating quote from that document's own text; it is
not a judgement of the filer, a claim about its present-day terms, or a conclusion this paper draws. Full filing texts are stored as
retrieved and \emph{unredacted}, and this is counted rather than assumed: 758 of the 1{,}123 plain-text
documents (67.5\%) carry an \texttt{/s/} signature block with a name, so most name at least one officer,
director or counsel. No redaction was applied and none is claimed; a released
\texttt{PERSONAL\_DATA.md} carries the full accounting, including a verified scan showing that no
derived artifact aggregates or indexes any of those names. Those documents are
already public at \texttt{sec.gov}, so this is a redistribution question rather than a disclosure
one.

\paragraph{Language and jurisdiction.} English only; United States only. The instruments and
terminology are those of U.S.\ venture financing under U.S.\ securities law, and no claim is made
that findings transfer to another jurisdiction's instruments without re-annotation.

\paragraph{Licensing.} \textbf{The licence under which the retrieved filing text is redistributed is
unresolved.} Three things the repository establishes: \texttt{LICENSE} is the MIT License and by its
own text covers software; no data licence, licence header or redistribution statement for the corpus
exists anywhere in it; and an SEC filing is authored by the filer and hosted by a government system,
so government-works rules do not automatically apply. We draw no legal conclusion in either
direction, and \textbf{we do not assert a redistribution right for the retrieved filing texts.} A
re-user can rely on the extracts, labels, quotes, audit outputs and code under the licences the
repository states; the full texts carry no asserted licence. The release includes all 1{,}145 of them,
so the question is open over material already distributed rather than a decision still pending. Every
table and figure here is reproducible from the extracts, labels and committed audit output alone; the
full texts are needed only to re-run the provenance audit's source-document check.

\paragraph{Intended use and out of scope.} Probity is an instrument for measuring answer
instability. It is not legal advice, not a compliance tool, and not a basis for a financing
decision: its central finding is that these models return different answers to the same question
about the same document, so acting on any single answer is what the benchmark shows to be unsafe. It
is not a general legal-reasoning benchmark, and at 470 items it is an instrument rather than a
training set. Tuning on it and reporting the result as capability would be misuse. A high score
means stability, which is not correctness.

%% file: tables/table3_categories.tex
\begin{table*}[t]
\centering
\scriptsize
\setlength{\tabcolsep}{4pt}
\begin{tabular}{lcccc}
\toprule
Model & cap table (1) & convertibles (12) & exit waterfall (1) & founder equity (5) \\
\midrule
gemma4-31b & 0.000 [.000,.324] & 0.000 [.000,.039] & 0.000 [.000,.658] & 0.000 [.000,.077] \\
haiku-4.5 & 0.000 [.000,.324] & 0.000 [.000,.040] & 0.000 [.000,.658] & 0.000 [.000,.077] \\
llama3.3-70b & 0.000 [.000,.324] & 0.000 [.000,.039] & 0.000 [.000,.658] & 0.022 [.004,.113] \\
gemini3-flash & 0.000 [.000,.324] & 0.000 [.000,.039] & 0.000 [.000,.658] & 0.000 [.000,.077] \\
mistral-large & 0.000 [.000,.324] & 0.000 [.000,.039] & 0.000 [.000,.658] & 0.065 [.022,.175] \\
gpt5-mini & 0.000 [.000,.324] & 0.000 [.000,.039] & 0.000 [.000,.658] & 0.043 [.012,.145] \\
deepseek-v4p & 0.000 [.000,.324] & 0.074 [.037,.146] & 0.000 [.000,.658] & 0.043 [.012,.145] \\
gpt-oss-120b & 0.000 [.000,.324] & 0.021 [.006,.074] & 0.000 [.000,.658] & 0.043 [.012,.145] \\
deepseek-v4f & 0.000 [.000,.324] & 0.042 [.016,.103] & 0.000 [.000,.658] & 0.022 [.004,.113] \\
minimax-m2.5 & 0.000 [.000,.324] & 0.042 [.016,.103] & 0.500 [.095,.905] & 0.065 [.022,.175] \\
gemma3-1b-qat & 0.375 [.137,.694] & 0.305 [.222,.404] & 1.000 [.342,1.000] & 0.261 [.156,.403] \\
gemma3-1b & 1.000 [.676,1.000] & 0.298 [.215,.397] & 1.000 [.342,1.000] & 0.348 [.227,.492] \\
\bottomrule
\end{tabular}

\vspace{4pt}

\setlength{\tabcolsep}{4pt}
\begin{tabular}{lcccc}
\toprule
Model & priced equity (16) & regulatory (5) & rights governance (7) & risk flag (5) \\
\midrule
gemma4-31b & 0.007 [.001,.041] & 0.038 [.007,.189] & 0.051 [.020,.125] & 0.000 [.000,.094] \\
haiku-4.5 & 0.015 [.004,.052] & 0.038 [.007,.189] & 0.026 [.007,.089] & 0.027 [.005,.138] \\
llama3.3-70b & 0.015 [.004,.052] & 0.077 [.021,.241] & 0.038 [.013,.107] & 0.000 [.000,.094] \\
gemini3-flash & 0.044 [.021,.094] & 0.000 [.000,.129] & 0.013 [.002,.069] & 0.054 [.015,.177] \\
mistral-large & 0.022 [.008,.063] & 0.077 [.021,.241] & 0.038 [.013,.107] & 0.027 [.005,.138] \\
gpt5-mini & 0.037 [.016,.084] & 0.154 [.061,.335] & 0.115 [.062,.205] & 0.000 [.000,.094] \\
deepseek-v4p & 0.030 [.012,.074] & 0.077 [.021,.241] & 0.051 [.020,.125] & 0.027 [.005,.138] \\
gpt-oss-120b & 0.037 [.016,.084] & 0.115 [.040,.290] & 0.090 [.044,.174] & 0.027 [.005,.138] \\
deepseek-v4f & 0.052 [.025,.103] & 0.115 [.040,.290] & 0.064 [.028,.141] & 0.081 [.028,.213] \\
minimax-m2.5 & 0.037 [.016,.084] & 0.308 [.165,.500] & 0.103 [.053,.190] & 0.000 [.000,.094] \\
gemma3-1b-qat & 0.296 [.226,.378] & 0.462 [.288,.645] & 0.256 [.173,.363] & 0.378 [.241,.539] \\
gemma3-1b & 0.467 [.385,.551] & 0.615 [.425,.776] & 0.256 [.173,.363] & 0.243 [.134,.401] \\
\bottomrule
\end{tabular}
\caption{Wobble by configuration and clause category on the audited corpus, each rate followed by its Wilson 95\% interval with leading zeros dropped. The eight categories are split across two panels; every panel lists all twelve configurations, ordered by aggregate wobble. Parenthesised numbers are the tasks each category retains after the provenance exclusion. \emph{cap table} and \emph{exit waterfall} fall below the three-task estimability floor and are shown for completeness only: no worst-category claim in this paper uses them. Intervals overlap almost everywhere, which is why no ordering between categories is asserted in the text.}
\label{tab:categories}
\end{table*}

%% file: tables/table5_audit.tex
\begin{table}[htbp]
\centering
\footnotesize
\setlength{\tabcolsep}{3.5pt}
\begin{tabular}{l r r}
\toprule
Quantity & Incl.\ flagged & Audited \\
\midrule
Items & 463 & 427 \\
Tasks & 60 & 52 \\
Model-item pairs & 5548 & 5116 \\
Tasks with non-zero wobble & 57 & 50 \\
Split-half reliability & 0.904 & 0.903 \\
Share at $p=1$ & 0.846 & 0.860 \\
Share in $0.3 \le p \le 0.7$ & 0.020 & 0.019 \\
Structural ceiling, acc.\ $\ge 0.9$ & 0.115 & 0.115 \\
Transfer $k=5$, median & 0.600 & 0.400 \\
Transfer $k=5$, chance & 0.175 & 0.200 \\
Transfer $k=10$, median & 0.500 & 0.400 \\
Wobble-accuracy $\rho$ & -0.603 & -0.487 \\
\bottomrule
\end{tabular}
\caption{Every headline quantity under both readings. The provenance audit flags 36 of 470 items. Both columns drop the 7 items removed at author adjudication: \emph{incl.\ flagged} keeps the 36 flagged items and \emph{audited} excludes them, so their item rows read 463 and 427. Pooled over all twelve configurations, wobble on flagged items is 0.255 [0.216, 0.298] against 0.087 [0.080, 0.095] on the rest, a difference of 0.167 [0.128, 0.211], computed by Newcombe's method on the two Wilson intervals. The interval excludes zero, so the audited reading is the paper's headline and this table is the record of what that choice moved.}
\label{tab:audit}
\end{table}

%% file: tables/table6_repair.tex
\begin{table*}[t]
\centering
\footnotesize
\setlength{\tabcolsep}{5pt}
\begin{tabular}{l r r l l l}
\toprule
Task & Items & Cells & Original window & Repaired window & Change \\
\midrule
convert vs.\ preference & 2 & 24 & 0.000 [0.000, 0.138] & 0.083 [0.023, 0.258] & $-$0.083 [$-0.258$, $0.067$] \\
multi-round dilution & 5 & 60 & 0.167 [0.093, 0.280] & 0.200 [0.118, 0.318] & $-$0.033 [$-0.172$, $0.107$] \\
option pool shuffle & 3 & 36 & 0.611 [0.449, 0.752] & 0.306 [0.180, 0.469] & 0.306 [$0.075$, $0.494$] \\
\midrule
\textbf{Pooled} & 10 & 120 & 0.267 [0.196, 0.352] & 0.208 [0.145, 0.289] & 0.058 [$-0.049$, $0.165$] \\
\bottomrule
\end{tabular}
\caption{The re-windowing repair, 10 items each measured in both conditions by all twelve configurations (120 paired cells). The repaired window is cut to the same length as the original and centred on the validating quote, so the only difference is whether the evidence is inside it. Registered before the experiment ran: the repaired rate would fall below 0.161. It is 0.208 [0.145, 0.289], above that threshold, so the prediction fails. The pooled change is 0.058 [$-0.049$, $0.165$] and contains zero. Change is the original rate minus the repaired one, so a positive value is a fall in wobble and a negative value a rise; 2 of the 3 tasks are negative. One task of three moves; the per-task split is descriptive and was not the registered test.}
\label{tab:repair}
\end{table*}

%% file: tables/table7_tasklevel.tex
\begin{table*}[t]
\centering
\footnotesize
\begin{tabular}{l S[table-format=1.0] S[table-format=3.0] c S[table-format=1.3] S[table-format=1.3] S[table-format=2.2] r}
\toprule
Unit & {$k$} & {Units} & Trunc. & {Mean} & {Chance} & {Ratio} & Perm. $p$ \\
\midrule
items & 3 & 425 & 0/90 & 0.144 & 0.014 & 10.23 & $<$ 0.00050 \\
 & 4 & 425 & 18/90 & 0.161 & 0.019 & 9.01 & $<$ 0.00050 \\
 & 5 & 425 & 36/90 & 0.187 & 0.024 & 8.91 & $<$ 0.00050 \\
\midrule
tasks & 3 & 52 & 18/90 & 0.315 & 0.115 & 2.87 & $<$ 0.00050 \\
 & 4 & 52 & 27/90 & 0.331 & 0.154 & 2.42 & $<$ 0.00050 \\
 & 5 & 52 & 42/90 & 0.369 & 0.192 & 2.28 & $<$ 0.00050 \\
\bottomrule
\end{tabular}
\caption{Transfer at every $k$ under both ranking units, over the same 425 clean items grouped into 52 tasks and the same ten frontier configurations, 90 ordered pairs each. A task's score is the share of its items that flip. Trunc. counts pairs where one side had fewer than $k$ non-zero units, so its top-$k$ was cut short rather than padded. Chance is recomputed per pair from the effective $k$ after truncation, which is why the coarser unit carries a chance line an order of magnitude higher and a correspondingly smaller ratio at the same permutation $p$. $p$ is over 2{,}000 permutations that shuffle each configuration's values among its own positions.}
\label{tab:tasklevel}
\end{table*}

%% file: tables/table8_serving.tex
\begin{table*}[t]
\centering
\footnotesize
\setlength{\tabcolsep}{4pt}
\begin{tabular}{l S[table-format=2.0] S[table-format=1.3] S[table-format=1.3] S[table-format=2.0] S[table-format=1.3]}
\toprule
Pairing & {Pairs} & {Mean} & {Median} & {Cross-fam.} & {Mean} \\
\midrule
both directly served & 6 & 0.567 & 0.600 & 4 & 0.550 \\
one of each & 42 & 0.167 & 0.200 & 42 & 0.167 \\
both routed & 42 & 0.152 & 0.000 & 42 & 0.152 \\
\bottomrule
\end{tabular}
\caption{Transfer at $k = 5$ by how each side of a pair is served, on the clean population. The two rightmost columns repeat the split after the two same-family configurations are removed, which is the check that the directly served result is not one model agreeing with its own sibling. Six ordered pairs from three configurations is the narrowest result in the paper and is reported as underpowered wherever it appears.}
\label{tab:serving}
\end{table*}

%% file: tables/table9_loo.tex
\begin{table}[t]
\centering
\footnotesize
\setlength{\tabcolsep}{4pt}
\begin{tabular}{l S[table-format=2.0] S[table-format=1.3] S[table-format=3.0] S[table-format=2.2]}
\toprule
Configuration dropped & {Left} & {Mean} & {Hits} & {Ratio} \\
\midrule
full lineup & 10 & 0.187 & 84 & 8.91 \\
\midrule
gemini3-flash & 9 & 0.197 & 71 & 9.43 \\
haiku-4.5 & 9 & 0.169 & 61 & 7.81 \\
gemma4-31b & 9 & 0.186 & 67 & 8.58 \\
mistral-large & 9 & 0.211 & 76 & 10.22 \\
llama3.3-70b & 9 & 0.197 & 71 & 9.31 \\
deepseek-v4p & 9 & 0.164 & 59 & 7.94 \\
gpt5-mini & 9 & 0.172 & 62 & 8.34 \\
deepseek-v4f & 9 & 0.175 & 63 & 8.47 \\
gpt-oss-120b & 9 & 0.194 & 70 & 9.41 \\
minimax-m2.5 & 9 & 0.200 & 72 & 9.68 \\
\bottomrule
\end{tabular}
\caption{Transfer at $k = 5$ with each frontier configuration dropped in turn, on the clean population. Left is how many configurations remain, Hits the total shared items over the ordered pairs among them, and Ratio those hits against the chance expectation recomputed for that lineup. The first row above the rule is the full lineup, so every row below it is read against that one.}
\label{tab:loo}
\end{table}

%% file: tables/table10_flags.tex
\begin{table*}[t]
\centering
\footnotesize
\setlength{\tabcolsep}{4pt}
\begin{tabular}{l l l c S[table-format=2.0] c l}
\toprule
Task & Oracle & Models say & Runs & {Floor} & Unan. & Adjudication \\
\midrule
preference\_seniority & pari-passu & stacked & 235/237 & 17 & yes & excluded \\
preference\_seniority & pari-passu & stacked & 224/226 & 5 & yes & excluded \\
participation\_type & participating & capped & 228/237 & 13 & yes & excluded \\
liquidation\_preference\_multiple & 1x & other & 220/229 & 11 & yes & excluded \\
flag\_uncapped\_participation & yes & no & 220/234 & 14 & no & excluded \\
flag\_uncapped\_participation & yes & no & 220/236 & 16 & no & excluded \\
price\_per\_share & 0.2 & 2.0 & 220/239 & 10 & no & excluded \\
\bottomrule
\end{tabular}
\caption{Items where the configurations' modal answers agree with each other and contradict the stored oracle label, ranked by run support. Of the 7, 7 have been adjudicated against the source filing. Runs is the number of the responses across all twelve configurations backing the models' answer. Floor is the weakest single configuration's own support out of 20 runs, carried as a column rather than applied as a filter: requiring every configuration to hold its answer on at least 18 of 20 admits none of these, and a criterion that empties the table is a result about the criterion. Unan. marks the items where all twelve configurations produced the same normalised answer. A flag is a candidate for human review and never a correction; no label changes until the author has read the filing and upheld it.}
\label{tab:flags}
\end{table*}